\documentclass[default]{sn-jnl}

\usepackage{graphicx}%
\usepackage{multirow}%
\usepackage{amsmath,amssymb,amsfonts}%
\usepackage{amsthm}%
\usepackage{mathrsfs}%
\usepackage[title]{appendix}%
\usepackage{xcolor}%
\usepackage{textcomp}%
\usepackage{manyfoot}%
\usepackage{booktabs}%
\usepackage{algorithm}%
\usepackage{algorithmicx}%
\usepackage{algpseudocode}%
\usepackage{listings}%
\usepackage{subfig}%
\usepackage{cprotect}%
\usepackage{lipsum}%
\usepackage{float}
\usepackage{subcaption}
\begin{document}

\title[Enhanced Muscle and Fat Segmentation for CT-Based Body Composition Analysis: A Comparative Study]{Enhanced Muscle and Fat Segmentation for CT-Based Body Composition Analysis: ~~~~~~~~~~~~~~~ A Comparative Study}


\author*[1]{\fnm{Benjamin} \sur{Hou}} 

\author[1]{\fnm{Tejas Sudharshan} \sur{Mathai}} 

\author[1]{\fnm{Jianfei} \sur{Liu}} 

\author[2]{\fnm{Christopher} \sur{Parnell}} 

\author[1]{\fnm{Ronald M.} \sur{Summers}} 

\affil[1]{\orgdiv{National Institutes of Health (NIH) Clinical Center},
\orgaddress{\city{Bethesda} \state{MD}, \country{USA}}}

\affil[2]{\orgdiv{Walter Reed National Military Medical Center},
\orgaddress{\city{Bethesda} \state{MD}, \country{USA}}}


\abstract{

\textbf{Purpose:} Body composition measurements from routine abdominal CT can yield personalized risk assessments for asymptomatic and diseased patients. In particular, attenuation and volume measures of muscle and fat are associated with important clinical outcomes, such as cardiovascular events, fractures, and death. This study evaluates the reliability of an Internal tool for the segmentation of muscle and fat (subcutaneous and visceral) as compared to the well-established public TotalSegmentator tool. 

\textbf{Methods:} We assessed the tools across 900 CT series from the publicly available SAROS dataset, focusing on muscle, subcutaneous fat, and visceral fat. The Dice score was employed to assess accuracy in subcutaneous fat and muscle segmentation. Due to the lack of ground truth segmentations for visceral fat, Cohen's Kappa was utilized to assess segmentation agreement between the tools.

\textbf{Results:} Our Internal tool achieved a 3\% higher Dice (83.8 vs. 80.8) for subcutaneous fat and a 5\% improvement (87.6 vs. 83.2) for muscle segmentation respectively. A Wilcoxon signed-rank test revealed that our results were statistically different with \textit{p} $<$ 0.01. For visceral fat, the Cohen's kappa score of 0.856 indicated near-perfect agreement between the two tools. Our internal tool also showed very strong correlations for muscle volume (\textit{R}$^2$=0.99), muscle attenuation (\textit{R}$^2$=0.93), and subcutaneous fat volume (\textit{R}$^2$=0.99) with a moderate correlation for subcutaneous fat attenuation (\textit{R}$^2$=0.45). 

\textbf{Conclusion:} Our findings indicated that our Internal tool outperformed TotalSegmentator in measuring subcutaneous fat and muscle. The high Cohen's Kappa score for visceral fat suggests a reliable level of agreement between the two tools. These results demonstrate the potential of our tool in advancing the accuracy of body composition analysis.

}




\maketitle

\section{Introduction}
\label{sec_intro}

The assessment of body composition, particularly the accurate segmentation of soft tissues such as subcutaneous fat, visceral fat, and muscle, has become a critical component in diagnostic imaging \cite{Lee2023_bodyCompositionAnalysis,doi:10.1148/radiol.2018181432}. Advances in computed tomography (CT) imaging have not only facilitated detailed body composition analysis, but also play a pivotal role in a range of medical applications, from disease characterization to surgical planning and radiation therapies \cite{makrogiannis2022multi, 10.1093/jncimonographs/lgad001}. This advancement in imaging technology demonstrates potential for enhanced `incidental' screening and tailored risk evaluation, benefiting both asymptomatic individuals and patients with existing medical conditions. For instance, the distribution and volume of visceral fat are closely linked to metabolic disorders and cardiovascular diseases, making their assessment crucial for early intervention strategies \cite{gruzdeva2018localization}. Similarly, understanding the balance between muscle and fat tissues is essential in evaluating nutritional status, which is particularly relevant in conditions like obesity, sarcopenia, and cachexia \cite{holmes2021utility, ozen2022association,Lee2023_bodyCompositionAnalysis}. In sports medicine and rehabilitation, analyzing muscle and fat distribution is crucial for creating personalized training and recovery programs \cite{wackerhage2021personalized}.

\begin{figure}[!ht]
    \centering
    \includegraphics[width=3.6cm]{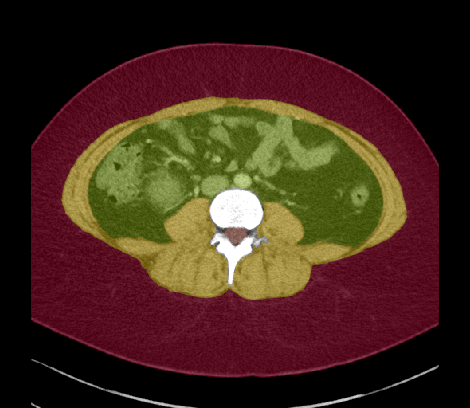}
    \includegraphics[width=3.6cm]{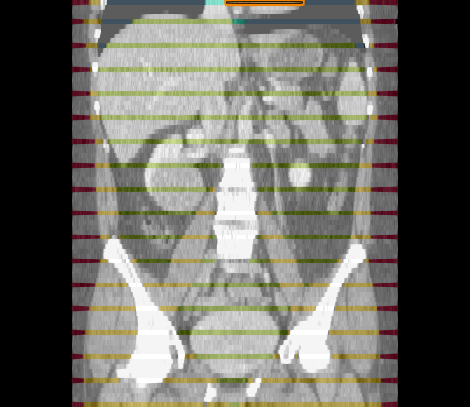}
    \includegraphics[width=3.6cm]{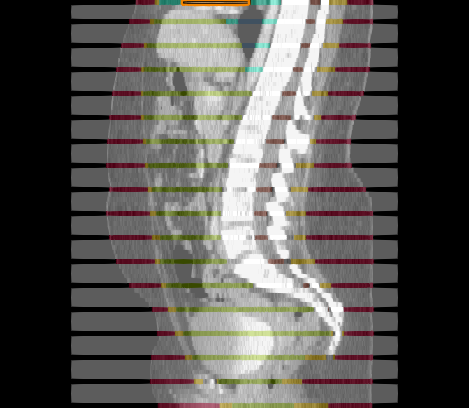}
    \cprotect\caption{Example axial, coronal and sagittal slice of \verb|case-042| from SAROS dataset. In the axial slice, the muscle (yellow), subcutaneous fat (red) and abdominal cavity (green) are shown. The grey regions in the coronal/sagittal views indicate no segmentation masks available in that area, while the streaks in between them contain segmentations. }
    \label{fig:saros-case-000}
\end{figure}

Similarly, in clinical research, such data significantly enhance our understanding of various health conditions and aid in the development of innovative treatments. This knowledge is particularly invaluable in oncology, where it plays a key role in tailoring treatment plans and monitoring the impact of therapies that can significantly alter body composition. Moreover, in surgical planning, especially in reconstructive or plastic surgery, the precise imaging of these tissues is essential for ensuring better outcomes and guiding post-operative care \cite{perrin2021effects, wopat2023body, aleixo2023association}. Recent developments in automated segmentation tools, such as TotalSegmentator \cite{doi:10.1148/ryai.230024}, have shown promising results in enhancing the efficiency and accuracy of these analyses. However, generalized tools in medical imaging, while versatile and broadly applicable, often do not perform with the same level of precision and efficacy as tools that are specifically targeted or tailored to particular tasks or conditions. The effectiveness of such tools compared to specialized solutions remains a subject of ongoing research.

In this study, we compare the effectiveness of the public TotalSegmentator tool against an internally developed tool for the task of muscle and fat (subcutaneous and visceral) segmentation in CT. We hypothesized that the internal tool developed specifically for muscle and fat segmentation would fare better than TotalSegmentator. Through experiments on the public SAROS dataset, we show that the internal tool fares better at the segmentation tasks, with statistical results to corroborate our findings. Our tool has substantial potential to be used for a broad range of clinical applications and offers opportunities for personalized risk assessment for patients.  

\section{Materials and Methods}
\label{sec_methods}

\subsection{Patient Population}

This study utilized deidentified data that is publicly available, thereby obviating the need for IRB approval. The dataset employed, known as the Sparsely Annotated Region and Organ Segmentation (SAROS) \cite{doi:10.25737/SZ96-ZG60, doi:10.1007/s10278-013-9622-7}, comprised of 900 CT series from 882 patients, evenly divided between 450 women and 450 men. These series were randomly selected from various TCIA \cite{doi:10.1007/s10278-013-9622-7} collections. 

The dataset contains CT volumes of 5mm slice thickness, with annotations provided in NIfTI format. These annotations covers 13 semantic body regions across 6 distinct body parts. The initial generation of annotations was carried out using body composition analysis tools developed by Koitka et al. \cite{DBLP:journals/corr/abs-2002-10776}, and subsequently reviewed and corrected by medical residents and students on every fifth axial slice, as illustrated in Fig. \ref{fig:saros-case-000}. Slices that were not reviewed were marked with an `ignore' label of value 255. In this retrospective study, we focused our analysis on three types of soft tissues: subcutaneous fat, visceral fat, and muscle. The SAROS dataset includes annotations for 13 semantic body regions and 6 body parts. However, ground truth segmentation labels within this dataset are only available for subcutaneous fat and muscle. Consequently, our analysis was limited to utilizing only the subcutaneous fat and muscle labels, with all other labels disregarded.

\subsection{TotalSegmentator}

TotalSegmentator \cite{doi:10.1148/ryai.230024} is a publicly accessible tool designed for segmenting over 117 distinct classes in CT images. It is apt for various applications, including organ volumetry, disease characterization, and planning for surgical or radiation therapy. This tool was developed using a training set of 1,204 CT examinations, encompassing a diverse array of scanners, institutions, and protocols to ensure its versatility and robustness in different clinical settings. Subcutaneous fat, skeletal muscle, and visceral fat structures fall under a separate task called 'tissue\_types', which, while publicly accessible, is subject to a non-commercial license agreement.

\subsection{Internal Tool}

Our internal tool leverages the 3D nnU-Net model \cite{Isensee21}, which is widely recognized and acclaimed as the \textit{de facto} standard in supervised segmentation. The training data was acquired using a 2D dual-branch network, as described in Liu et al. \cite{Liu23}. This 2D dual-branch network was initially developed to alleviate the extensive and time-consuming annotation burden associated with full CT volumes, enabling the generation of precise segmentations of muscle and fat across all slices of a CT scan. 

The dual-branch network features a shared encoder and two duplicate decoders. It was trained using a combination of a few strongly labeled and a large number of weakly labeled datasets; the strongly labeled data included manual annotations of muscle, visceral fat, and subcutaneous fat on each CT slice. The weak labels, generated automatically via a level-set method \cite{Burns20}, were prone to segmentation errors. The dual-branch network was trained through a mixed supervision approach utilizing both strong and weak labels. Throughout the training process, the weakly labeled data was periodically refined by the strong decoder in a self-supervised manner. Upon completion of the dual-branch network's training, it was applied to all CT volumes to generate dense annotations across all CT series. These annotations were then utilized as training data for the 3D full-resolution nnU-Net. 

\subsection{Statistical Analysis} 

As previously mentioned, this retrospective study focuses on three types of soft tissues: subcutaneous fat, visceral fat, and muscle. While both TotalSegmentator (TS) and our Internal tool are capable of segmenting all three tissue types, the SAROS dataset only includes ground truth labels for subcutaneous fat and muscle. After the Internal tool and TotalSegmentator were executed on the CT series in the dataset, the Dice coefficient was utilized to compute the similarity between the predicted segmentations and the ground-truth annotations. Since not all slices in the dataset were labeled, Dice score calculation was confined to the ``valid'' regions of interest, which were delineated by the body mask provided. For all analyses, slices lacking labels, as well as background pixels, were excluded. This approach ensures that our evaluation focused solely on the relevant anatomical areas.

After assessing the normality of the Dice score distribution, a Wilcoxon signed-rank test was employed to determine any statistical differences. Due to the absence of ground truth labels for visceral fat in the dataset, Cohen's Kappa \cite{Cohen1960_kappaScore} was used to evaluate the agreement between TotalSegmentator and our internal tool in segmenting visceral fat. Cohen's Kappa is a statistical measure that captures the agreement between two raters, taking into account the possibility of agreement occurring by chance. In addition, graphs correlating the ground-truth segmentations contrasted against the predictions were also plotted with overlaid $R^2$ values. Bland-Altman analysis was also conducted through the calculation of volume differences (biases) and averages for each structure to determine agreement. The Dice and kappa scores were calculated using the Scikit-learn library (Version 1.3.1) in Python (Version 3.9.10). All statistical tests were performed using RStudio (Version 2023.06.1+524). 

\section{Results}
\label{sec_results}

Our study's focus is on comparing the performance of different tools, rather than comparing different scans or patients. Each tool is applied to measure the same scan, with the expectation that the reported volume of tissue types by each tool should be consistent. Should our comparison have been between scans or patients, standardizing the area of measurement would indeed be necessary, such as constraining to the abdomen section (featuring structures L1-L5 and T9-T12) only. In contrast, had our comparison focused on scans or patients, it would have necessitated standardizing the area of measurement. For instance, such standardization could involve limiting the measurement to specific sections of the body, such as the abdomen area, which includes anatomical structures between L1-L5 and T9-T12. 

\begin{figure}[!ht]
\centering
\subfloat[][]{\includegraphics[width=5cm]{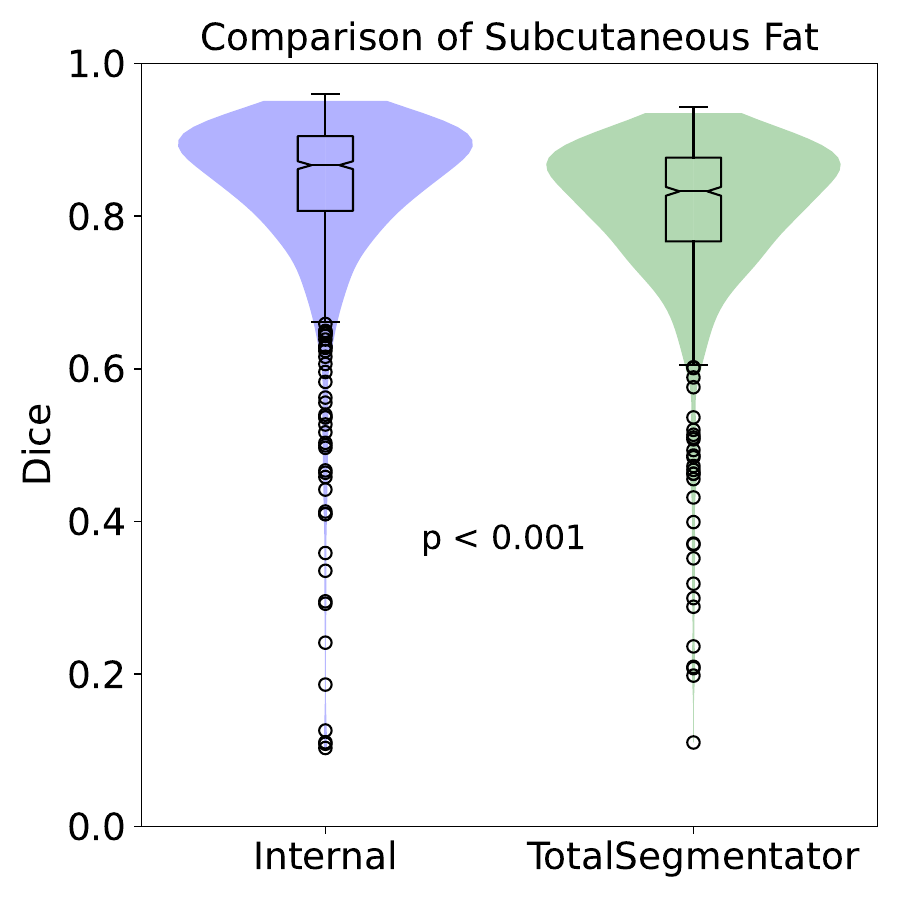}}
\subfloat[][]{\includegraphics[width=5cm]{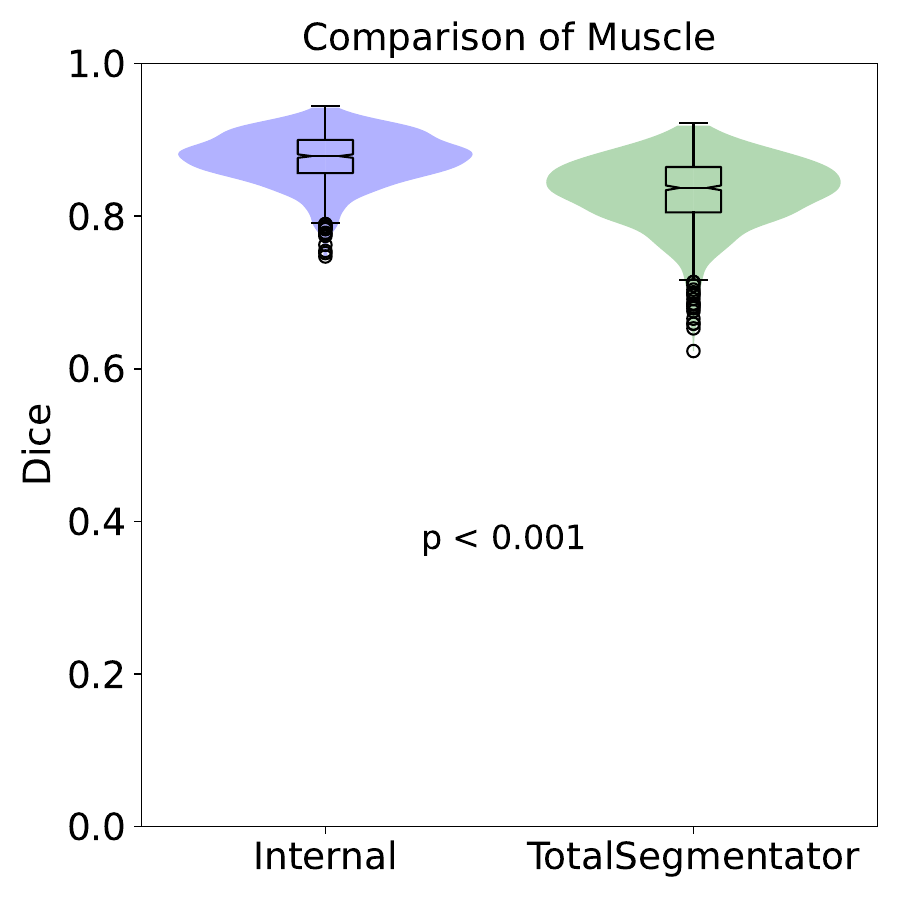}}
\caption{Violin plot of TotalSegmentator (green) vs. our internal tool (blue) for the segmentation of (a) subcutaneous fat and (b) muscle.}
\label{fig:box_plots}
\end{figure}

Table \ref{tab:results_dice} presents a direct comparison of the segmentation capabilities of TotalSegmentator and our Internal tool, specifically focusing on subcutaneous fat and muscle segmentation. Fig. \ref{fig:box_plots} shows violin plots to visualize the distributions of Dice Scores for both TotalSegmentator and Internal tool. Dice scores in Table \ref{tab:results_dice} are presented as means $\pm$ standard deviation, along with the 25th and 75th percentiles (IQR), for both subcutaneous fat and muscle. For subcutaneous fat, TotalSegmentator achieved an average score of 80.8 (±10.4) with an IQR range of [76.7, 87.7]. In contrast, our Internal tool demonstrated a slightly higher mean Dice Score of 83.8 (±10.9) with an IQR range of [80.7, 90.5]. With respect to muscle, TotalSegmentator attained a mean score of 83.2 (±4.6) and [80.5, 86.4] IQR. In contrast, our Internal tool outperformed it by 5\% as a mean score of 87.6 (±3.3) and [85.6, 90] IQR was obtained. Notably, as depicted in Fig. \ref{fig:box_plots}, the internal tool exhibits fewer outliers compared to TotalSegmentator, particularly in muscle segmentation, indicating a more consistent and reliable performance. These results suggest that while both tools are effective for soft tissue segmentation, the Internal tool was superior in segmenting both subcutaneous fat and muscle with $p < 0.01$.

\begin{table}[!ht]
\centering
\captionsetup{width=.9\textwidth}
\caption{Table of Dice scores: TotalSegmentator vs. Internal Tool for subcutaneous fat and muscle Segmentation. Scores are shown with mean, standard deviation, and Inter-Quartile Ranges (IQR).}
\begin{tabular}{@{}lcc@{}}
\toprule
                    & Subcutaneous Fat  & Muscle ~\\ 
\midrule
~TotalSegmentator   & 80.8 $\pm$ 10.4 [76.7, 87.7] 
                    & 83.2 $\pm$ 4.6 [80.5, 86.4] ~\\
~Internal           & 83.8 $\pm$ 10.9 [80.7, 90.5] 
                    & 87.6 $\pm$ 3.3 [85.6, 90.0] ~\\ 
\bottomrule
\end{tabular}
\label{tab:results_dice}
\end{table}

\begin{table}[!ht]
\centering
\captionsetup{width=.9\textwidth}
\caption{Cohen's Kappa scores: Agreement of TotalSegmentator and Internal tool for segmentation of subcutaneous fat, visceral fat, and muscle. Scores are shown with mean, standard deviation, and Inter-Quartile Ranges (IQR).}
\begin{tabular}{@{}lc@{}}
\toprule
                    & Cohen's Kappa ~\\ 
\midrule
~Subcutaneous Fat   & 0.874 $\pm$ 0.066 [0.854, 0.913]  ~\\ 
~Visceral Fat       & 0.856 $\pm$ 0.074 [0.830, 0.906]  ~\\ 
~Muscle             & 0.837 $\pm$ 0.033 [0.819, 0.861]  ~\\ 
\bottomrule
\end{tabular}
\label{tab:results_cohens_kappa}
\end{table}

\begin{figure}[!ht]
    \centering
    \subfloat[][TS Muscle Vol]{\includegraphics[width=0.245\linewidth]{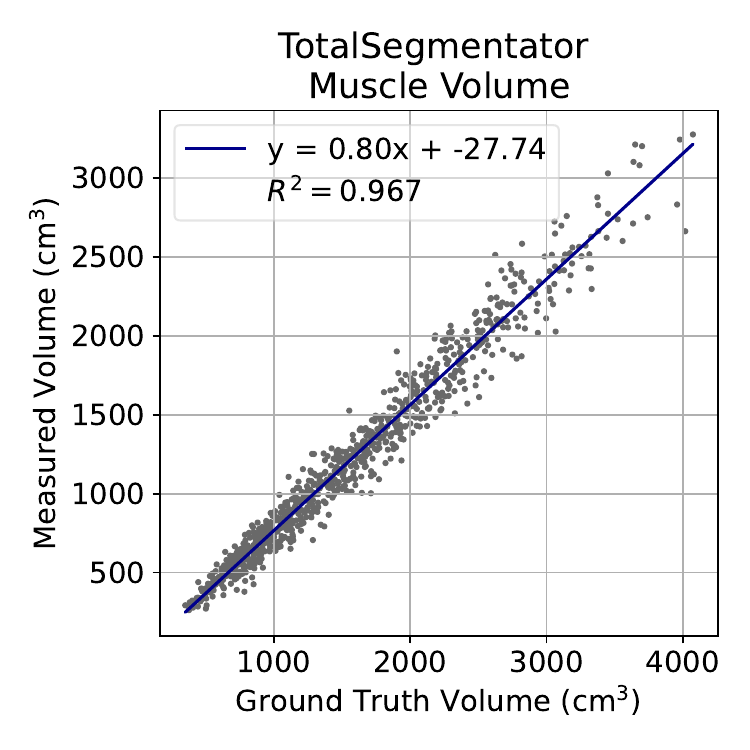}}
    \subfloat[][TS Muscle Att]{\includegraphics[width=0.245\linewidth]{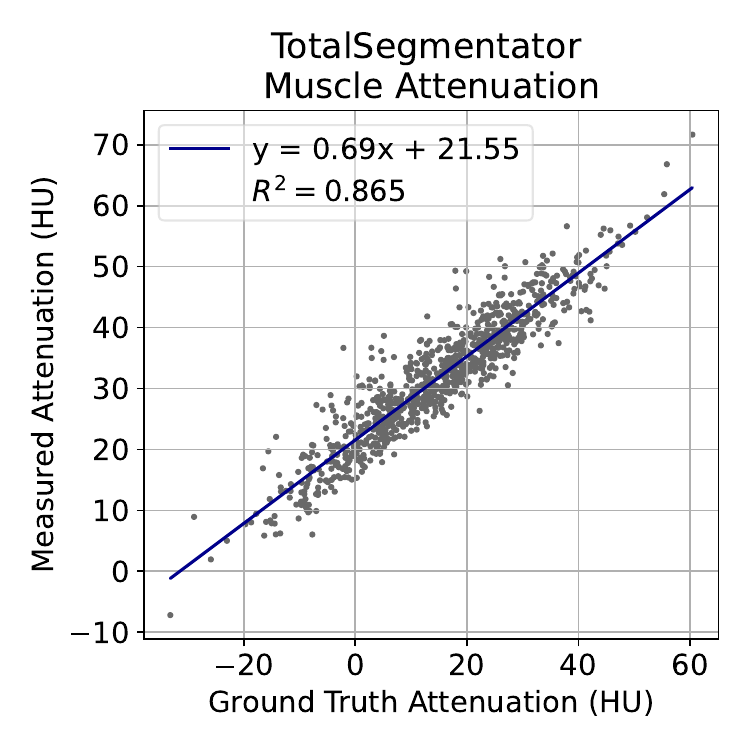}}
    \subfloat[][TS Fat Vol]{\includegraphics[width=0.245\linewidth]{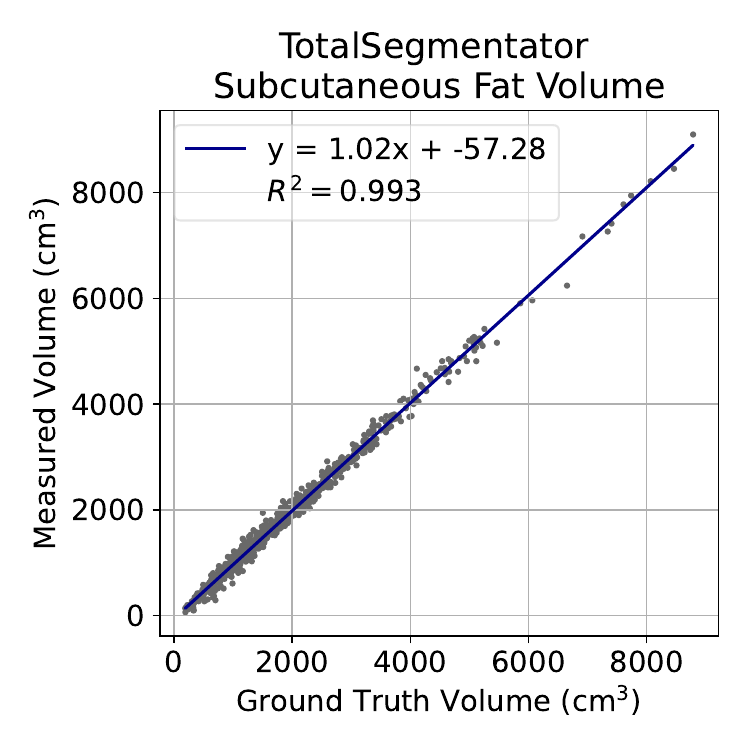}}
    \subfloat[][TS Fat Att]{\includegraphics[width=0.245\linewidth]{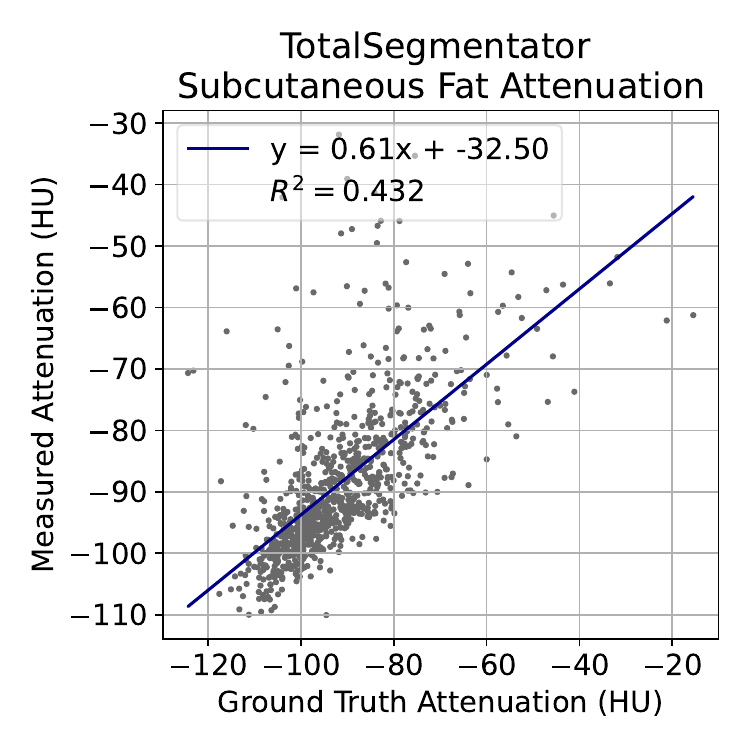}} \\ 
    \subfloat[][Int Muscle Vol]{\includegraphics[width=0.245\linewidth]{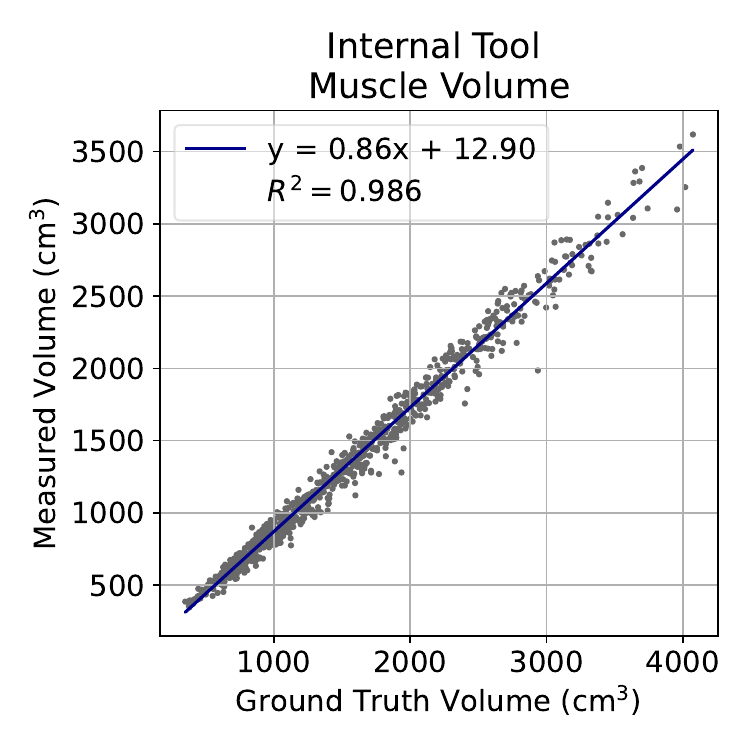}}
    \subfloat[][Int Muscle Att]{\includegraphics[width=0.245\linewidth]{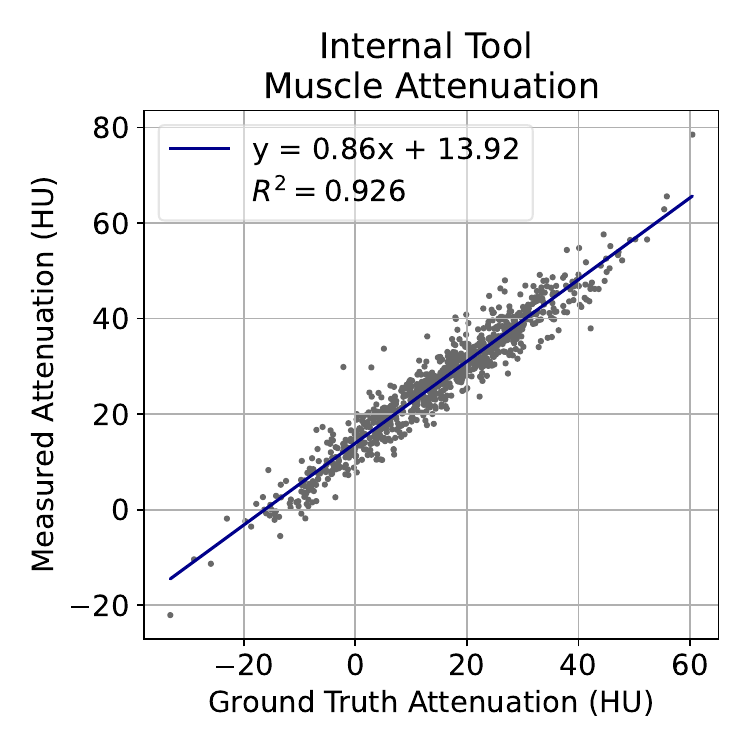}}
    \subfloat[][Int Fat Vol]{\includegraphics[width=0.245\linewidth]{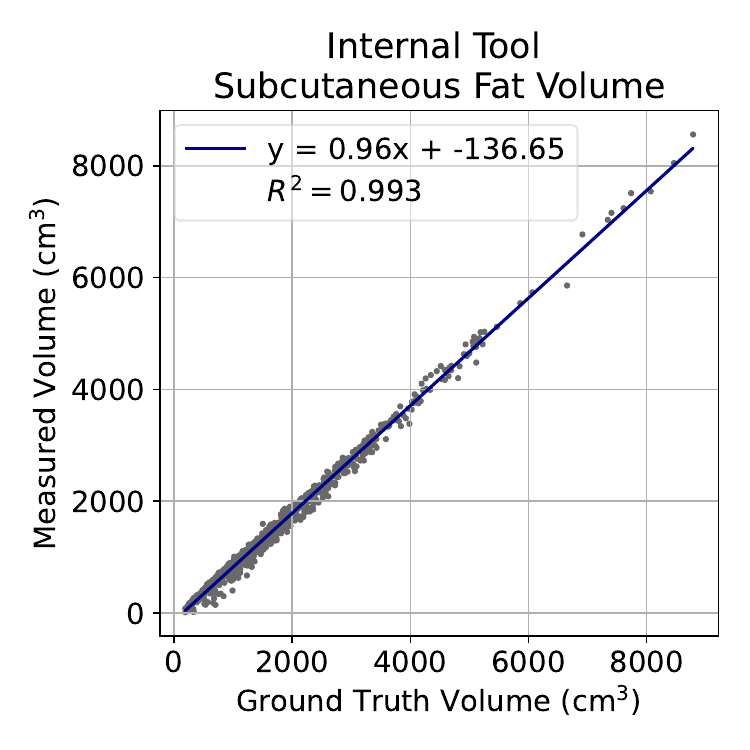}}
    \subfloat[][Int Fat Att]{\includegraphics[width=0.245\linewidth]{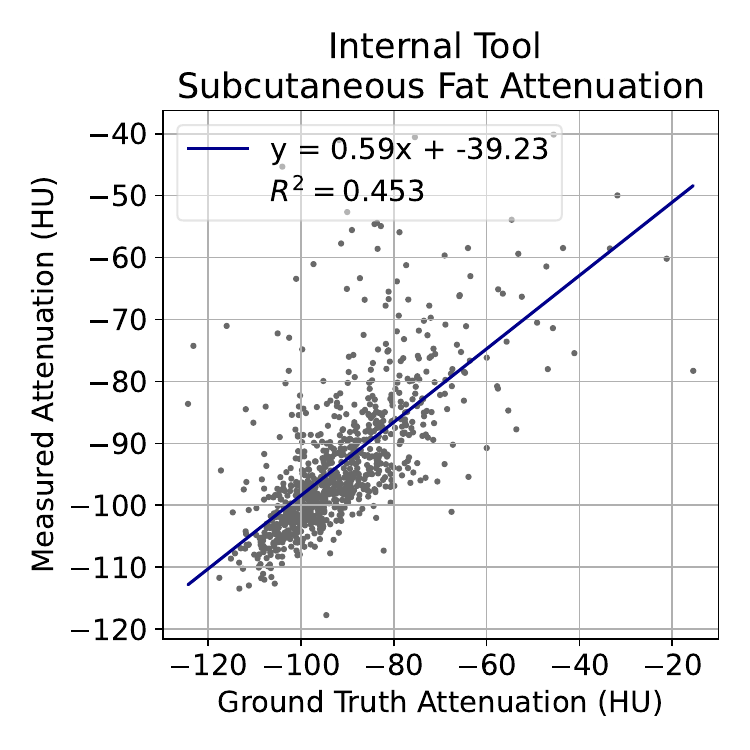}}    
    \caption{$R^2$ correlation plots of the automatic segmentation results compared against ground truth annotations. Top Row: TotalSegmentator (TS). Bottom Row: Internal (Int) tool. L-to-R: Muscle Volume, Muscle Attenuation, Fat Volume, Fat Attenuation.}
    \label{fig:r2-plots}
\end{figure}

SAROS provides ground truth labels on every fifth axial slice, but these labels are limited to muscle and subcutaneous fat only. Given the absence of labels for visceral fat, the entire CT volume was utilized for comparisons between TotalSegmentator and our Internal tool. It's important to note that subcutaneous fat and visceral fat are considered separate structures and do not overlap. The kappa scores in Table \ref{tab:results_cohens_kappa} indicated a high level of concordance between the two tools across all three tissue types. Fig. \ref{fig:r2-plots} shows $R^2$ correlation plots for the volume and attenuation of the different structures respectively. The average Hounsfield Unit (HU) of muscle attenuation for both TotalSegmentator and our Internal tool exhibit a strong correlation with $R^2$ values of 0.87 and 0.93, respectively, with our Internal tool outperforming it by a notable margin. This is supported by the similarly strong correlation observed with muscle volume, yielding $R^2$ values of 0.97 and 0.99, respectively. For subcutaneous fat, despite a significant uncertainty in the average HU values for both tools, with $R^2$ values of 0.43 for TotalSegmentator and 0.45 for our Internal tool. Nevertheless, the region was accurately segmented, with fat volume estimation showing a high correlation, evidenced by an $R^2$ value of 0.99 for both tools.

\begin{figure}[!ht]
\centering
\subfloat[][TS Muscle]{\includegraphics[width=0.245\linewidth]{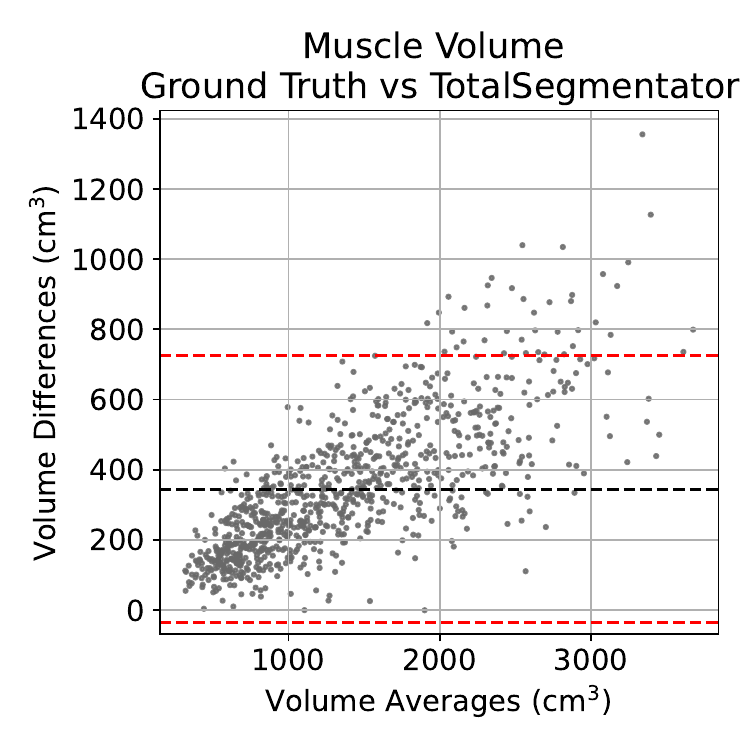}}
\subfloat[][Internal Muscle]{\includegraphics[width=0.245\linewidth]{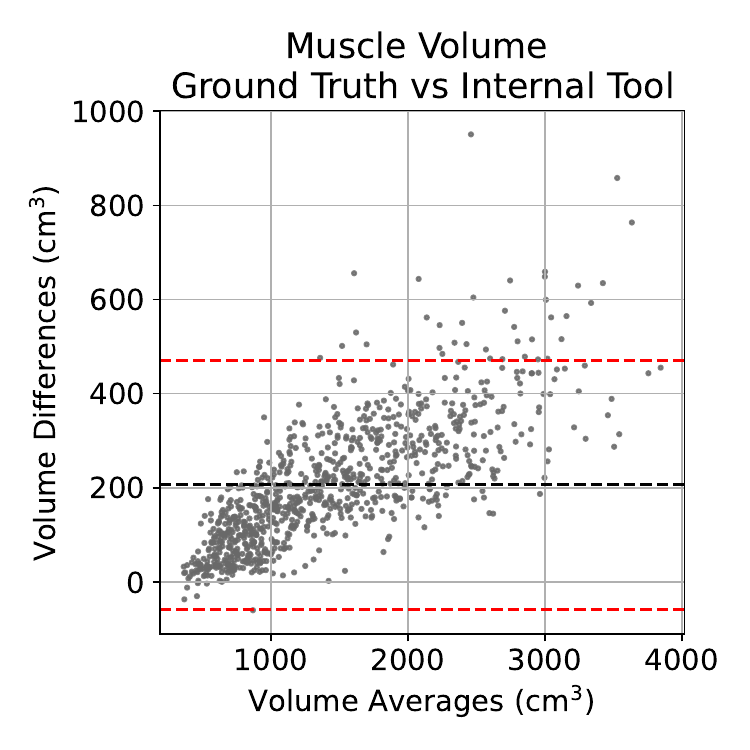}}
\subfloat[][TS Fat]{\includegraphics[width=0.245\linewidth]{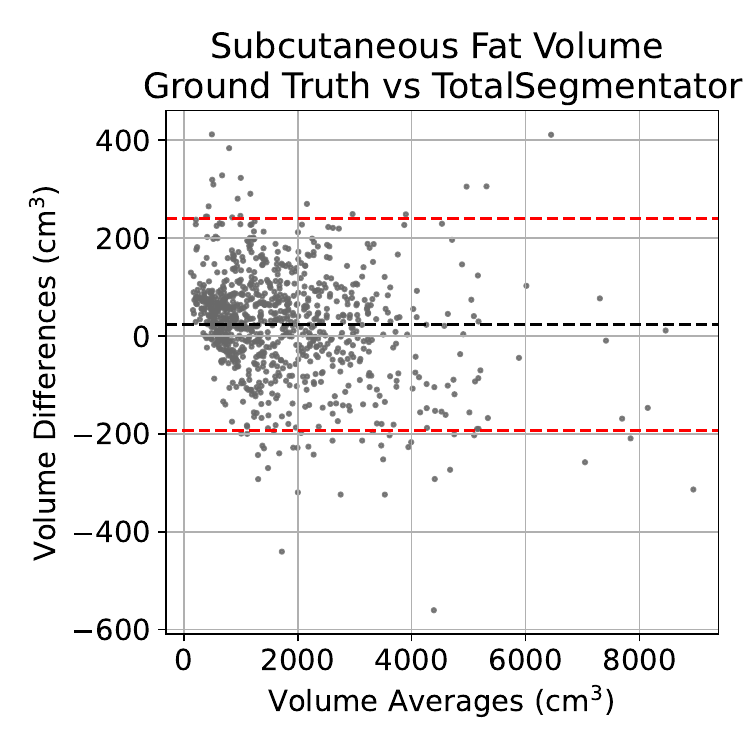}}
\subfloat[][Internal Fat]{\includegraphics[width=0.245\linewidth]{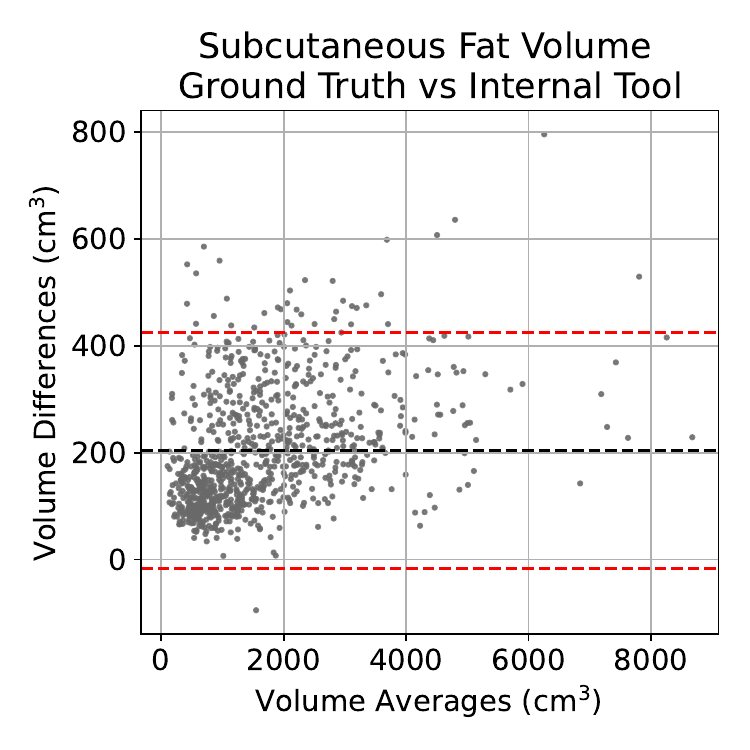}}
\caption{Bland-Altman plots of the volume measurements between the automatic segmentations compared against manual annotations. L-to-R: TotalSegmentator Muscle Volume, Internal Muscle Volume, TotalSegmentator Subcutaneous Fat, Internal Subcutaneous Fat.}
\label{fig:ba-plots}
\end{figure}

Fig. \ref{fig:ba-plots} displays Bland-Altman plots for muscle and subcutaneous fat volume estimation of the tools compared to the manual annotations. For both tools measuring muscle volume, there's a noticeable positive skew in the data. The Internal tool demonstrated a significantly lower bias, approximately 250cm$^3$, in comparison to the TotalSegmentator, which exhibited a bias around 500cm$^3$. For the subcutaneous fat volume estimation, there is a distinct concentration of data points on the left-hand side. The Internal tool has a slight positive skew also with a higher bias at around +200cm$^3$ compared to TotalSegmentator that is around 0cm$^3$.

\begin{figure}[!ht]
    \centering
    \includegraphics[width=3cm]{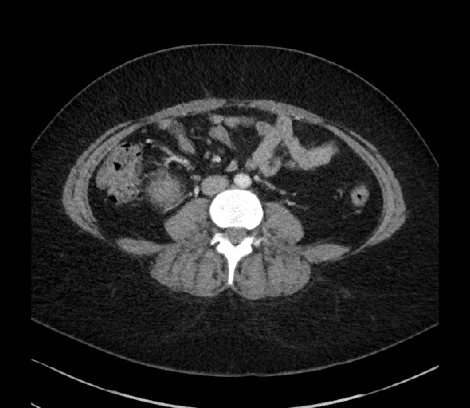}
    \includegraphics[width=3cm]{gt_case042_axial_40.png}
    \includegraphics[width=3cm]{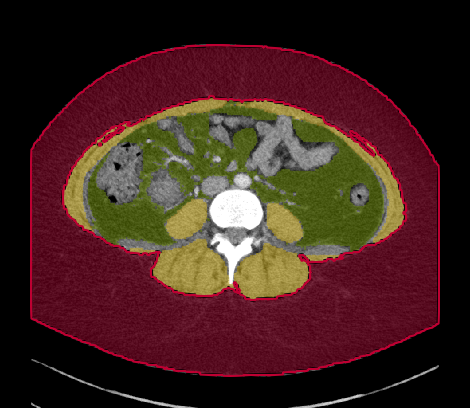}
    \includegraphics[width=3cm]{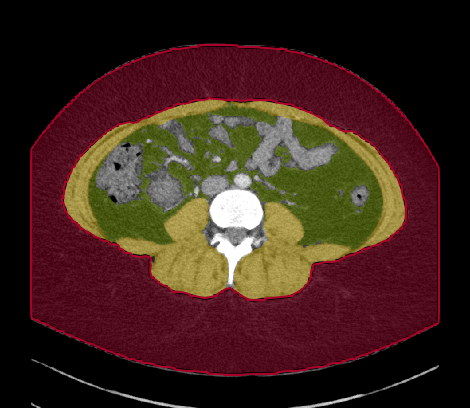}\\ 
    \includegraphics[width=3cm]{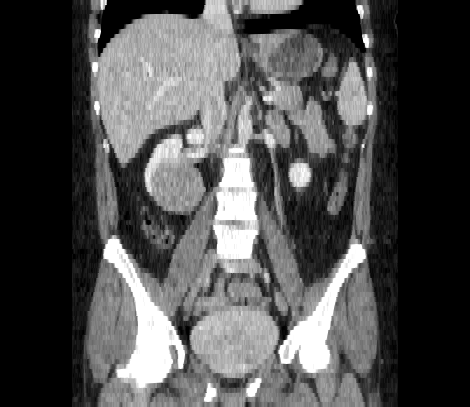}
    \includegraphics[width=3cm]{gt_case042_coronal_255.png}
    \includegraphics[width=3cm]{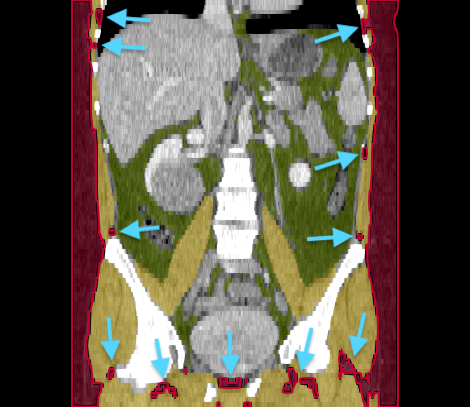}
    \includegraphics[width=3cm]{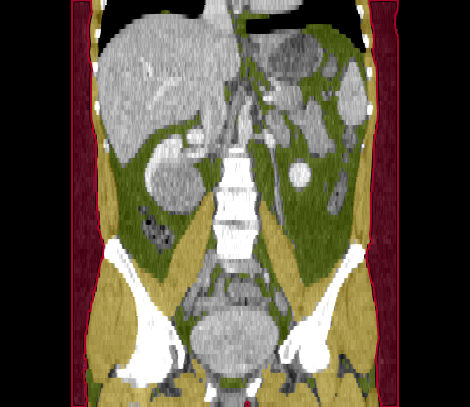}\\
    \includegraphics[width=3cm]{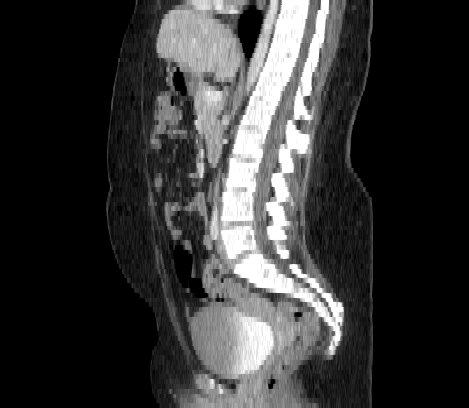}
    \includegraphics[width=3cm]{gt_case042_sagittal_256.png}
    \includegraphics[width=3cm]{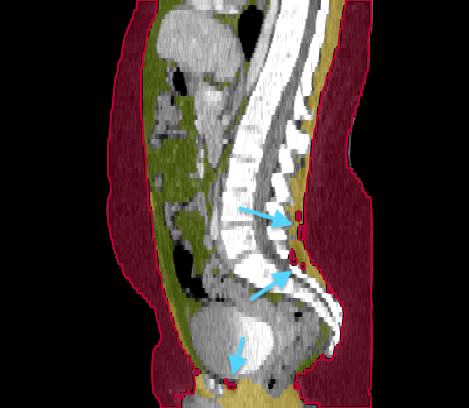}
    \includegraphics[width=3cm]{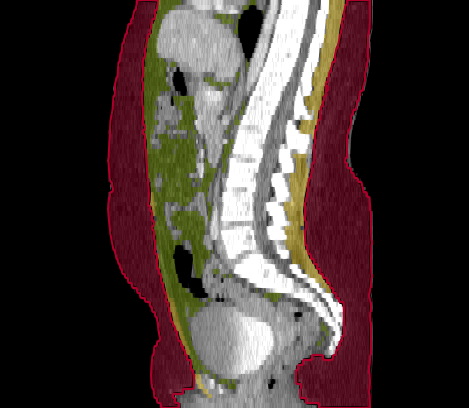}
    \cprotect\caption{Example segmentation of \verb|case-042|. Top-to-Bottom: axial, coronal, sagittal views. L-to-R: CT image, manual annotation (ground truth), TotalSegmentator segmentation, Internal tool segmentation. Red: Subcutaneous Fat, Yellow: Muscle, Green: Internal Abdominal Cavity (ground truth only) / Visceral Fat, Grey: No ground truth labels. Blue arrows shows over segmentation of subcutaneous fat by TotalSegmentator where it was correctly segmented as muscle by our Internal tool.}
    \label{fig:case042}
\end{figure}

Fig. \ref{fig:case042} shows and example segmentation of body composition by TotalSegmentator and our Internal tool. In a comparison of segmentation accuracy, our internal tool outperformed TotalSegmentator, achieving Dice scores of 0.947 for Subcutaneous Fat and 0.884 for Muscle, compared to TotalSegmentator's scores of 0.919 and 0.809, respectively. Additionally, our internal tool exhibited a robust Cohen's Kappa score of 0.876, further demonstrating its strong agreement compared to a popular and widely used tool. TotalSegmentator has shown a tendency to over-segment subcutaneous fat, as indicated by the blue arrows in Fig. \ref{fig:case042}. This is particularly evident in areas such as the muscle between the ribs and within the pelvic cavities.

\begin{figure}
    \centering
    \begin{minipage}[c]{0.02\textwidth}
        \centering
        \rotatebox{90}{\small\textsc{case531}}
    \end{minipage}
    \begin{minipage}[c]{0.68\textwidth}
        \centering
        \includegraphics[height=2cm,width=2cm]{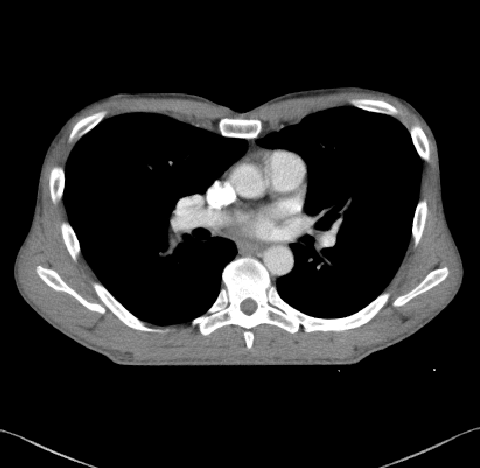}     
        \includegraphics[height=2cm,width=2cm]{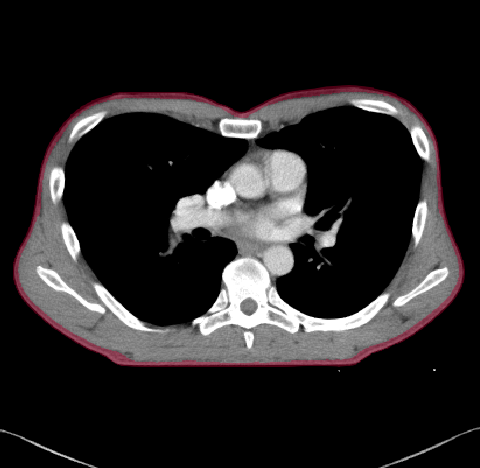}     
        \includegraphics[height=2cm,width=2cm]{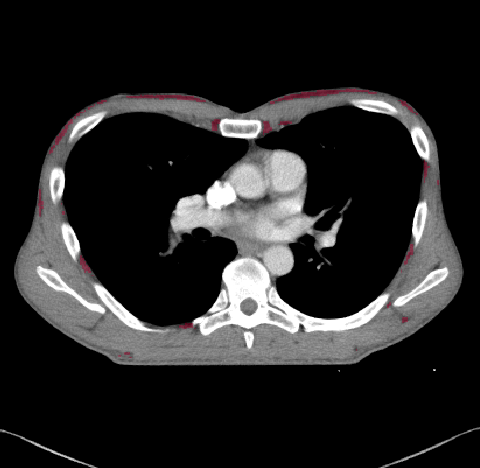}     
        \includegraphics[height=2cm,width=2cm]{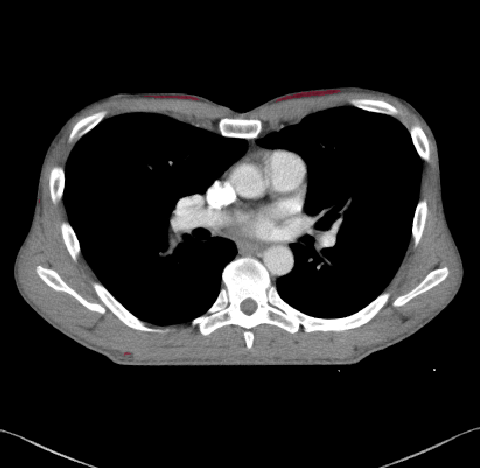} 
    \end{minipage}
    \begin{minipage}[c]{0.2\textwidth}
        \footnotesize
        TS Dice: 0.209\\
        Int Dice: 0.111\\
        TS-Int Kappa: 0.345
    \end{minipage}

        \begin{minipage}[c]{0.02\textwidth}
        \centering
        \rotatebox{90}{\small\textsc{case547}}
    \end{minipage}
    \begin{minipage}[c]{0.68\textwidth}
        \centering
        \includegraphics[height=2cm,width=2cm]{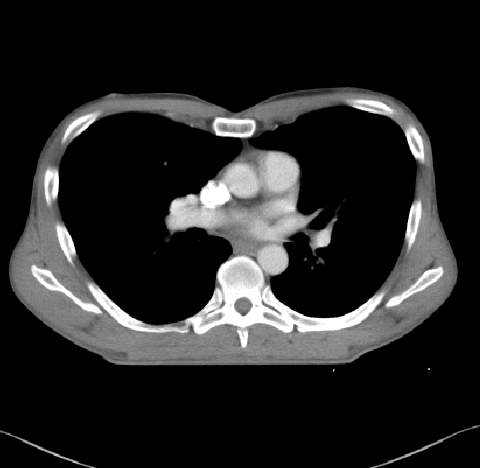}     
        \includegraphics[height=2cm,width=2cm]{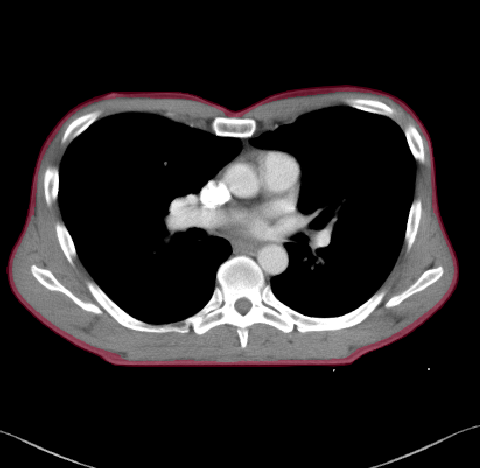}     
        \includegraphics[height=2cm,width=2cm]{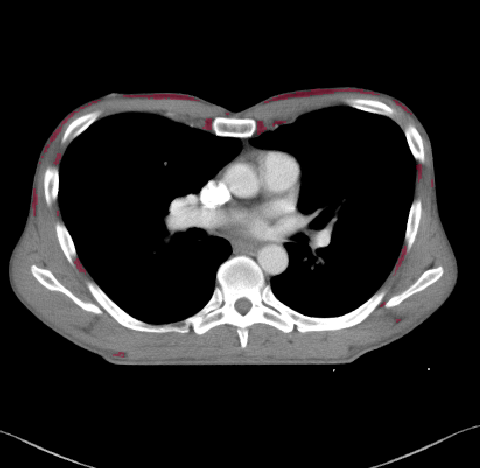}     
        \includegraphics[height=2cm,width=2cm]{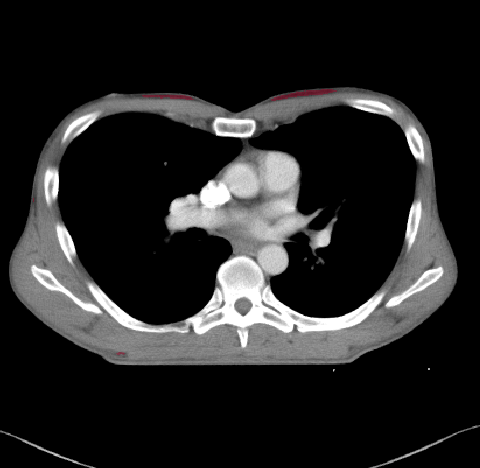} 
    \end{minipage}
    \begin{minipage}[c]{0.2\textwidth}
        \footnotesize
        TS Dice: 0.208\\
        Int Dice: 0.109\\
        TS-Int Kappa: 0.340 
    \end{minipage}

        \begin{minipage}[c]{0.02\textwidth}
        \centering
        \rotatebox{90}{\small\textsc{case866}}
    \end{minipage}
    \begin{minipage}[c]{0.68\textwidth}
        \centering
        \includegraphics[height=2cm,width=2cm]{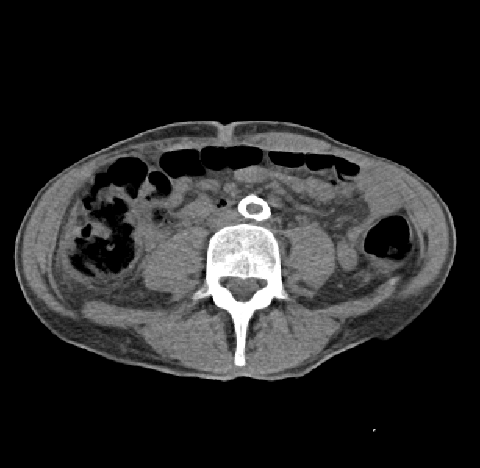}     
        \includegraphics[height=2cm,width=2cm]{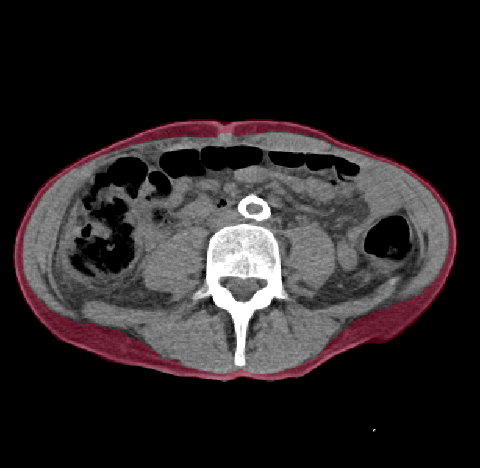}     
        \includegraphics[height=2cm,width=2cm]{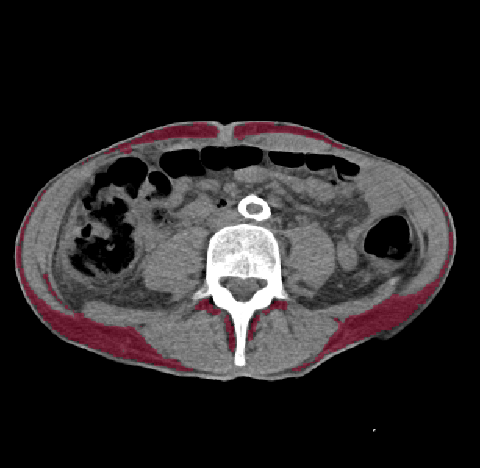}     
        \includegraphics[height=2cm,width=2cm]{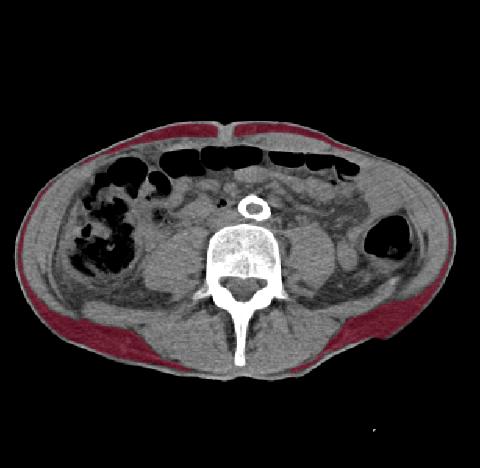} 
    \end{minipage}
    \begin{minipage}[c]{0.2\textwidth}
        \footnotesize
        TS Dice: 0.462\\
        Int Dice: 0.462\\
        TS-Int Kappa: 0.750 
    \end{minipage}

    \cprotect\caption{Comparison of subcutaneous fat segmentation failure cases by TotalSegmentator and Internal tool. Top-to-bottom: \verb|case-531|, \verb|case-547|, \verb|case-886| from SAROS dataset. L-to-R: CT image, manual annotation (ground truth), TotalSegmentator segmentation, Internal tool segmentation.}
    \label{fig:fail-cases}
\end{figure}

TotalSegmentator and the Internal Tool demonstrate a high level of segmentation agreement, as evidenced by the Cohen's Kappa scores presented in Table \ref{tab:results_cohens_kappa}. Figure \ref{fig:box_plots} reveals that both tools perform effectively in segmenting muscle tissue, achieving Dice coefficients greater than 0.6, however, this level of performance does not extend to the segmentation of subcutaneous fat. Most instances of segmentation failure (Dice scores $<$ 0.5) occur in patients with a low body fat percentage. This issue is compounded by the imaging resolution; even at 1mm, it hinders the clear delineation of subcutaneous fat, which is situated between the skin (dermal layers) and muscle, often covering only a few pixels. The observed low Dice coefficients are attributed to the coarse annotations provided by the annotators, rather than to the segmentation tools themselves as shown in Fig. \ref{fig:fail-cases}.

\section{Discussion and Conclusion}
\label{sec_discussion}

Through our experiments, the Internal tool achieved a 3\% higher Dice (83.8 vs. 80.8) for subcutaneous fat and a 5\% improvement (87.6 vs. 83.2) for muscle segmentation respectively. The results yielded by the internal tool were statistically different \textit{p} $<$ 0.01. However, from the $R^2$ correlation plots in Fig. \ref{fig:r2-plots} for subcutaneous fat, a significant uncertainty was seen in the average HU values for both tools: 0.43 for TotalSegmentator and 0.45 for our Internal tool.

The considerable standard deviation in HU values within the subcutaneous fat layer can be attributed to its diverse composition. This layer, primarily composed of adipocytes, also contains fibroblasts, blood vessels, nerve cells, lymphatic vessels, immune cells, hair follicles, and sweat glands, each with differing densities. These varying densities result in a broad spectrum of HU values, as captured in CT scans. The contrast between the low-density adipocytes and the higher-density components within the layer leads to the observed variability in HU readings. 

The variability can also be attributed to several other factors: the quality and noise in CT images affecting segmentation precision, limitations in the segmentation algorithm especially if not tailored for subcutaneous fat, variability in fat composition and density, and the choice of thresholding in segmentation. This complexity not only highlights the multifaceted nature of the subcutaneous layer, but also underscores the challenge in accurately segmenting and analyzing it using CT imaging. Despite the variations in HU, the subcutaneous fat volume demonstrated a high correlation for both tools with an $R^2$ value of 0.99, indicating accurate segmentation of the subcutaneous fat region.

The skewness in the Bland-Altman plots in Fig. \ref{fig:ba-plots} suggests a tendency for the differences between the two methods under comparison to increase as the magnitude of the measurement decreases. Such a distribution pattern indicates a potential systematic bias in the measurements, particularly at lower values. For the concentration of data points on the left-hand side in the subcutaneous fat volume estimation, the pattern indicates that the agreement between the two methods being compared is more consistent at lower measurement values. Such a concentration suggests that for smaller magnitudes of the variable being measured, the two methods yield closer results, implying better concordance in this range. However, this also raises questions about the performance of the methods at higher values, as the relative sparsity of data points on the right-hand side of the plot may indicate a divergence in the methods' readings or a limitation in the range of data sampled. 

Furthermore, segmenting muscle tissue is a relatively easier task due to its clearly defined visual boundaries. In contrast, the delineation of fat can be challenging, as its boundaries are not always distinct. This challenge stems from the fact that fat and water-rich tissues (such as specific soft tissues) can exhibit similar Hounsfield Units (HUs), complicating their differentiation. Fat typically has a slightly negative HU value, often in the range of -50 to -100 HU, whereas water has an HU of 0. However, the HU values of soft tissues can range from -10 to +60 HU, depending on the specific tissue type and its water content. 

The overlapping HU values between fat and certain soft tissues create a significant challenge for differentiation based solely on attenuation properties. This is particularly true for visceral fat, where the close proximity and interleaving of blood vessels, bowel, and organs give it a complex shape. Although fat and muscle have distinct HU values, the HU values of the bowel, vessels, and organs may closely resemble those of muscle, especially in non-contrast CT scans, or when the CT scan's resolution is too low to clearly differentiate between these tissue types. Furthermore, fat deposits can be located within muscle tissue, indicating that HU values are not the primary reason for the segmentation difficulty for visceral fat.

In summary, this study has demonstrated that our internal tool significantly outperforms the more generalized TotalSegmentator in accurately segmenting subcutaneous fat, visceral fat, and muscle in CT series. Our findings are supported by high Dice scores and strong correlations ($R^2$) with manual annotations, and is further corroborated by Bland-Altman plots demonstrating consistent agreement and minimal bias. The enhanced accuracy and consistency of our internal tool hold significant promise for a range of clinical applications, such as providing improved personalized risk assessments for patients at risk of adverse cardiovascular events and fractures.

\section{Acknowledgments}
\label{sec_Acknowledgments}

\noindent
\textbf{Funding}: This work was supported by the Intramural Research Program of the NIH Clinical Center (project number 1Z01 CL040004).

\noindent
\textbf{Ethical approval}: All procedures performed in studies involving human participants were in accordance with the ethical standards of the institutional and/or national research committee and the 1964 Helsinki declaration and its later amendments or comparable ethical standards. For this study, informed consent was not required.

\noindent
\textbf{Conflict of Interest}: RMS receives royalties from iCAD, MGB, Philips, PingAn, ScanMed, and Translation Holdings. His lab received research support from PingAn. The authors have no additional conflicts of interest to declare.


\clearpage

\bibliography{sn-bibliography}
\bibliographystyle{unsrt}

\end{document}